\documentclass[10pt,journal,compsoc]{IEEEtran}

\usepackage{amsmath,amssymb,amsthm}
\usepackage{algorithm,algorithmic}
\usepackage{kbordermatrix}
\usepackage{xcolor}
\usepackage{graphicx}
\usepackage{comment}
\usepackage[pagebackref]{hyperref}
\usepackage{pgfplots}
\pgfplotsset{compat=newest}
\usepackage{xfrac}

\ifCLASSOPTIONcompsoc
  \usepackage[nocompress]{cite}
\else
  \usepackage{cite}
\fi

\ifCLASSINFOpdf
\else
\fi

\newcommand{\wrt}{w.r.t.\ }
\newcommand{\ie}{i.e.,\ }
\newcommand{\eg}{e.g.,\ }
\newcommand{\cf}{cf.\ }
\newcommand{\etal}{et al.\ }
\newcommand{\tm}{\!\!\times\!\!}

\newcommand{\cln}{\,:\,}

\newcommand{\templateTest}{\textsc{TemplateTest}}
\newcommand{\templateFinder}{\textsc{TemplateFinder}}
\newcommand{\templateReduction}{\textsc{TemplateReduction}}
\newcommand{\ve}[1]{v(#1)}
\newcommand{\cam}[1]{{\mathcal{#1}}}
\renewcommand{\arraystretch}{1.1}

\newcommand{\hide}[1]{}

\theoremstyle{plain}

\theoremstyle{definition}
\newtheorem{example}{Example}
\newtheorem{definition}{Definition}
\theoremstyle{remark}

\begin{document}

\title{Automatic Solver Generator for Systems of Laurent Polynomial Equations}

\author{Evgeniy~Martyushev,
        Snehal~Bhayani,
        and~Tomas~Pajdla
\IEEEcompsocitemizethanks{
\IEEEcompsocthanksitem E. Martyushev is with the Department of Mathematical Analysis and Mathematics Education, Institute of Natural Sciences and Mathematics, South Ural State University, Chelyabinsk, \hide{Lenin avenue 76, }Russia.\protect\\
E-mail: martiushevev@susu.ru
\IEEEcompsocthanksitem S. Bhayani is with the Center for Machine Vision and Signal Analysis, University of Oulu, Finland\protect\\
E-mail: snehal.bhayani@oulu.fi
\IEEEcompsocthanksitem T. Pajdla is with the Czech Institute of Informatics, Robotics, and Cybernetics, Czech Technical University in Prague, Prague 6, Czech Republic\protect\\
E-mail: pajdla@cvut.cz
}
}

\IEEEtitleabstractindextext{%
\begin{abstract}
In computer vision applications, the following problem often arises: Given a family of (Laurent) polynomial systems with the same monomial structure but varying coefficients, find a solver that computes solutions for any family member as fast as possible. Under appropriate genericity assumptions, the dimension and degree of the respective polynomial ideal remain unchanged for each particular system in the same family. The state-of-the-art approach to solving such problems is based on elimination templates, which are the coefficient (Macaulay) matrices that encode the transformation from the initial polynomials to the polynomials needed to construct the action matrix. Knowing an action matrix, the solutions of the system are computed from its eigenvectors. The important property of an elimination template is that it applies to all polynomial systems in the family. In this paper, we propose a new practical algorithm that checks whether a given set of Laurent polynomials is sufficient to construct an elimination template. Based on this algorithm, we propose an automatic solver generator for systems of Laurent polynomial equations. The new generator is simple and fast; it applies to ideals with positive-dimensional components; it allows one to uncover partial $p$-fold symmetries automatically. We test our generator on various minimal problems, mostly in geometric computer vision. The speed of the generated solvers exceeds the state-of-the-art in most cases. In particular, we propose the solvers for the following problems: optimal 3-view triangulation, semi-generalized hybrid pose estimation and minimal time-of-arrival self-calibration. The experiments on synthetic scenes show that our solvers are numerically accurate and either comparable to or significantly faster than the state-of-the-art solvers.
\end{abstract}

}

\maketitle

\IEEEdisplaynontitleabstractindextext

%
\IEEEpeerreviewmaketitle

\ifCLASSOPTIONcompsoc
\IEEEraisesectionheading{\section{Introduction}}
\else
\section{Introduction}
\label{sec:intro}
\fi

\IEEEPARstart{M}{any}
problems of applied science can be reduced to finding common roots of a system of multivariate (Laurent) polynomial equations. Such problems arise in chemistry, mathematical biology, theory of ODE's, geodesy, robotics, kinematics, acoustics, geometric computer vision, and many other areas. For some problems, it is only required to find all (or some) roots of a particular polynomial system, and the root-finding time does not matter much.

In contrast, other problems require finding roots for a family of polynomial systems with the same monomial structure, but different coefficient values.
For a given set of coefficients, the roots must be found quickly and with acceptable accuracy. Under appropriate genericity assumptions on the coefficients, the dimension, and degree of the corresponding polynomial ideal remain unchanged. The state-of-the-art approach to solving such problems is to use symbolic-numeric solvers based on elimination templates~\cite{kukelova2008automatic,larsson2017efficient,bhayani2020sparse,martyushev2022optimizing}. These solvers have two main parts. In the first offline part, an elimination template is constructed. The template consists of a map (formulas) from input data to a (Macaulay) coefficient matrix. The structure of the template is the same for each set of coefficients. In the second online phase, the coefficient matrix is filled with the data of a particular problem, reduced by the Gauss--Jordan (G--J) elimination, and used to construct an eigenvalue/eigenvector computation problem of an action matrix that delivers the solutions of the system.

While the offline phase is not time critical, the online phase has to be computed very fast (usually in sub-milliseconds) to be useful for robust optimization based on the RANSAC schemes~\cite{fischler1981random}. The speed of the online phase is mainly determined by two operations, namely the G--J elimination of the template matrix and the eigenvalue/eigenvector computation of the action matrix.  Therefore, one approach to generating fast solvers, is to find elimination templates that are as small as possible. The size of the elimination templates affects not only the speed of the resulting solvers, but also their numerical stability. The latter is more subtle, but experiments show that the larger templates have worse stability without special stability enhancing techniques, see e.g.\  \cite{byrod2007improving,telen2018stabilized}.


\subsection{Contribution}

We propose a new automatic generator of elimination templates for efficiently solving systems of Laurent polynomial equations. The advantages of our generator are as follows.
\begin{itemize}
\item {\bf Flexibility:} It finds elimination templates for a possibly redundant number of roots. In some cases, it can significantly reduce the template size and thus speed up the root computation.
\item {\bf Versatility:} (i) It is applicable to polynomial ideals with positive-dimensional components; (ii) It is also applicable to uncovering the partial $p$-fold symmetries to generate smaller templates.
\item {\bf Simplicity:} By and large, it uses only manipulations with sets of monomials and G--J elimination on matrices over a finite field.
\end{itemize}

We demonstrate our method on a variety of minimal problems mostly in geometric computer vision. For many of them, we have constructed solvers that are faster than the state-of-the-art.

We propose a solver for the famous problem of optimal 3-view triangulation~\cite{stewenius2005hard,byrod2008column,larsson2017polynomial}, which is naturally formulated as a system of Laurent polynomial equations. Our solver for this problem is numerically accurate and slightly faster than the state-of-the-art solvers from~\cite{byrod2008column,larsson2017polynomial}.

We also propose a fast solver for the semi-generalized hybrid pose estimation problem~\cite{bhayani2023hybsemigenpose}. Defined as the problem of estimating the relative pose of a pinhole camera with unknown focal length \wrt a calibrated generalized camera, from a hybrid set of one 2D-2D and three 2D-3D point correspondences, its original formulation in~\cite{bhayani2023hybsemigenpose} used a homography-based formulation along with the elimination ideal method~\cite{kukelova2017clever}. However, this led to large expressions for the polynomial coefficients, resulting in slow solvers. In comparison, our solver relies on a depth-based formulation that results in a Laurent polynomial system. The coefficients of this system are much simpler expressions. Therefore, the solver generated using our proposed AG is $20-30$ times faster than the solvers based on the homography formulation. 

Finally, we propose solvers for the $4s/6r$ and $5s/5r$ Time-of-Arrival minimal problems~\cite{kuang2013complete,larsson2017polynomial,larsson2020upgrade}. Our solvers have comparable numerical accuracy and are $1.3-1.8$ times faster than the state-of-the-art solvers from~\cite{larsson2020upgrade}.

\subsection{Related work}

Elimination templates are matrices that encode the transformation from polynomials of the initial system to polynomials needed to construct the action matrix. Knowing an action matrix, the solutions of the system are computed from its eigenvectors. \emph{Automatic generator} (AG) is an algorithm that takes a polynomial system as input and outputs an elimination template for the action matrix computation.

\noindent\textbf{Automatic generators:} The first automatic generator was built in~\cite{kukelova2008automatic}, where the template was constructed iteratively by expanding the initial polynomials with their multiples of increasing degree. This AG has been widely used by the computer vision community to construct polynomial solvers for a variety of minimal problems, \eg~\cite{bujnak20093d,bujnak2010new,kukelova2013fast,zheng2015minimal,saurer2015minimal,nakano2015globally}, see also~\cite[Tab.~1]{larsson2017efficient}. Paper~\cite{larsson2017efficient} introduced a non-iterative AG based on tracing the Gr\"obner basis construction and subsequent syzygy-based reduction. This AG allowed fast template construction even for hard problems. An alternative AG based on the use of sparse resultants was proposed in~\cite{bhayani2020sparse}. This method, together with~\cite{larsson2018beyond}, are currently the state-of-the-art automatic template generators.

\noindent\textbf{Improving stability:} The standard way to construct the action matrix from a template requires performing its LU decomposition. For large templates, this operation often leads to significant round-off and truncation errors, and hence to numerical instability. The series of papers~\cite{byrod2007improving,byrod2008column,byrod2009fast} addressed this problem and proposed several methods of improving stability, e.g.\ by performing a QR decomposition with column pivoting on the step of constructing the action matrix from a template.

\noindent\textbf{Optimizing formulations:} Choosing an appropriate formulation of a minimal problem can drastically simplify finding its solutions. The paper~\cite{kukelova2017clever} proposed the variable elimination strategy, which reduces the number of unknowns in the initial polynomial system. For some problems, this strategy led to significantly smaller templates~\cite{larsson2017making,kileel2018distortion}.

\noindent\textbf{Optimizing templates:} Much effort has been spent on speeding up the action matrix method by optimizing the template construction step. The paper~\cite{naroditsky2011optimizing} introduced a method to optimize templates by removing some unnecessary rows and columns. The method in~\cite{kukelova2014singly} exploited the sparsity of elimination templates by converting a large sparse template into the so-called singly-bordered block-diagonal form. This allowed splitting the initial problem into several smaller subproblems that are easier to solve. In paper~\cite{larsson2018beyond}, the authors proposed two methods that significantly reduced the size of elimination templates. The first method used the so-called Gr\"obner fan of a polynomial ideal to construct templates \wrt all possible standard bases of the quotient space. The second method went beyond Gr\"obner bases and introduced a random sampling strategy to construct non-standard bases. In~\cite{martyushev2022optimizing}, the authors proposed a heuristic greedy optimization strategy to reduce the templates obtained by the non-iterative AG from~\cite{larsson2017efficient}.

\noindent\textbf{Optimizing root solving:} Complex roots are spurious for most problems arising in applications. The paper~\cite{bujnak2012making} introduced two methods to avoid the computation of complex roots, resulting in a significant speedup of polynomial solvers.

\noindent\textbf{Discovering symmetries:} Polynomial systems for certain minimal problems may have hidden symmetries. Uncovering these symmetries is another way to optimize templates. This approach was demonstrated for the simplest partial $p$-fold symmetries in~\cite{kuang2014partial,larsson2016uncovering}. A more general case was studied in~\cite{duff2022galois}. 

\noindent\textbf{Laurent polynomial ideals:} Some application problems can be naturally formulated as a system of Laurent polynomial equations, and only the toric roots of the system are of interest. Clearly, any Laurent polynomial equation can be transformed either into an ordinary polynomial equation by taking its numerator, or into a system of ordinary polynomial equations by introducing new variables. It follows that any AG for ordinary polynomials can be also applied to Laurent polynomials. However, such an approach can have unwanted consequences: increasing the number of variables, increasing the total degree of polynomials, introducing false (non-toric) roots. All this can complicate the root-finding process. Working directly in the Laurent polynomial ring is preferable as it provides more ``degrees of freedom'' in choosing action polynomial and constructing shifts of the initial polynomials. The Gr\"obner and the border bases for Laurent polynomial ideals were introduced in~\cite{pauer1999grobner} and~\cite{mourrain2014toric} respectively. An eigenvalue method for solving square systems of Laurent polynomial equations has been proposed in~\cite{telen2020numerical}. For Laurent systems with more polynomials than the number of variables, \ie non-square systems, a sparse resultant-based method has been proposed in~\cite{bhayani2020sparse} which uses Newton polytopes~\cite{cox2006using} to generate the elimination template as a resultant matrix.

\noindent\textbf{The most related work:} Our work is essentially based on the results of papers~\cite{byrod2009fast,kukelova2008automatic,larsson2018beyond,martyushev2022optimizing}.

\section{Solving sets of Laurent monomials}

We use $\mathbb K$ for a field, $X = \{x_1, \ldots, x_k\}$ for a set of $k$ variables, $R = \mathbb K[X, X^{-1}]$ for the $\mathbb K$-algebra of Laurent polynomials over $\mathbb K$.

Let $F = \{f_1, \ldots, f_s\} \subset R\setminus \mathbb K$ and $J = \langle F\rangle$ be the ideal generated by $F$. Let
\[
\mathcal V = \{p \in (\mathbb K \setminus \{0\})^k \cln f_1(p) = \ldots = f_s(p) = 0\}
\]
be the set of common roots of $F$. We assume that $\mathcal V$ is $0$-dimensional, \ie it is a finite set of points. More generally, $\mathcal V$ is reducible and one of its components is $0$-dimensional, \ie $\mathcal V = \widetilde{\mathcal V} \cup \mathcal V_0$ with $\dim \mathcal V_0 = 0$. The positive-dimensional variety $\widetilde{\mathcal V}$ consists of superfluous unfeasible roots. This case was addressed in~\cite{larsson2017polynomial} for polynomial systems. In the sequel, we assume that $\dim \mathcal V = 0$.

It is clear that there exists 
$
(\alpha_1^j, \ldots, \alpha_k^j) \in \mathbb Z^k_{\geq 0}
$
such that
\[
\widetilde f_j = x_1^{\alpha_1^j}\ldots x_k^{\alpha_k^j}f_j \in \mathbb K[X]
\]
for each $j = 1, \ldots, s$. Thus, $\mathcal V$ can be also obtained as a set of common roots of the polynomial system $\widetilde F = 0$, where $\widetilde F = \{\widetilde f_1, \ldots, \widetilde f_s\}$. However, the use of $\widetilde F$ instead of $F$ may result in the appearance of superfluous roots that do not belong to the torus $(\mathbb K \setminus \{0\})^k$. Saturating these roots is an additional non-trivial problem in general. Furthermore, the total degrees of the polynomials in $\widetilde F$ can increase significantly, which can lead to larger elimination templates. In contrast, our examples show that working directly with the Laurent polynomials leads to smaller elimination templates and thus to faster solvers, \cf Problems \#35 and \#36 in Tab.~\ref{tab:res} below.

We start by generalizing the definition of solving bases (in this paper we will use the term "solving sets") from~\cite{byrod2009fast} for Laurent polynomials. For simplicity, we restrict ourselves to the solving sets consisting of monomials. Let
\[
U = \{x_1^{\alpha_1}\ldots x_k^{\alpha_k} \cln (\alpha_1, \ldots, \alpha_k) \in \mathbb Z^k\}
\]
be the set of Laurent monomials in $X$.

We denote by $\ve{\mathcal A}$ the vector consisting of the elements of a finite set of Laurent monomials $\mathcal A \subset U$ which are ordered according to a certain total ordering on $U$, \eg the graded reverse lex ordering (grevlex) with $x_1 > \ldots > x_k$ which compares monomials first by their total degree, \ie $\alpha_1 + \ldots + \alpha_k$, and breaks ties by smallest degree in $x_k$, $x_{k - 1}$, etc. Note that grevlex is not a well-ordering on $U$, but this is of no importance for our purposes.

\begin{definition}
Let $\mathcal B \subset U$ and $a \in R\setminus \mathbb K$. Let us define the vector
\begin{equation}
\label{eq:vectorC}
C := a\, T_1 \ve{\mathcal B} - T_0 \ve{\mathcal B} \in R^d,
\end{equation}
where $d = \#\mathcal B$, $T_0, T_1 \in \mathbb K^{d \times d}$, and $\det T_1 \neq 0$. The set of monomials $\mathcal B$ is called the \emph{solving set} for the ideal $J$ if the following condition holds:
\begin{enumerate}
\item[(C1)] $C \subset J$, \ie each element of $C$ is a Laurent polynomial from $J$.
\end{enumerate}
In this case the polynomial $a$ is called the \emph{action polynomial}.
\end{definition}

If $\mathcal B$ is a solving set for $J$, then $C(p) = 0$ for any $p \in \mathcal V$ and hence we come to the generalized eigenproblem~\cite{golub2013matrix}
\begin{equation}
\label{eq:geig}
T_0 \ve{\mathcal B(p)} = a(p)\, T_1 \ve{\mathcal B(p)}.
\end{equation}
It follows that
\[
a(p) \in \sigma(T_0,T_1) = \{\lambda \in \mathbb K \cln \det(T_0 - \lambda T_1) = 0\}.
\]
In this paper we restrict ourselves to the case $\det T_1 \neq 0$, which guarantees that the set $\sigma(T_0,T_1)$ is finite~\cite{golub2013matrix}. Since the matrix $T_1$ is invertible, the problem~\eqref{eq:geig} can be solved as the regular eigenproblem for the action matrix $T_1^{-1}T_0$. The drawback of such an approach is that an ill-conditioned matrix $T_1$ can cause significant inaccuracies in the computed eigenvalues. On the other hand, there is a numerically backward stable QZ algorithm~\cite{kressner2005numerical} for solving the problem~\eqref{eq:geig}.

For each $p \in \mathcal V$ there exists $\lambda \in \sigma(T_0,T_1)$ such that $a(p) = \lambda$. If the related eigenspace $\ker(T_0 - \lambda T_1)$ is $1$-dimensional and $u$ is its basis vector, then $u = \ve{\mathcal B(p)}$ up to scale.

Note that the vector $C$ may vanish at a point $p \notin \mathcal V$. Therefore the set $\{a(p) \cln p \in \mathcal V\}$ may be a proper subset of $\sigma(T_0,T_1)$, \ie it may happen that $d > \#\mathcal V$. In this case, the solving set is said to be \emph{redundant}~\cite{byrod2009fast}. It may also happen that $d = \#\mathcal V$ or $d < \#\mathcal V$. The latter case applies e.g.\ to systems with the partial $p$-fold symmetries~\cite{kuang2014partial,larsson2016uncovering}.

Next, given a solving set $\mathcal B$ let us introduce the following additional condition:
\begin{enumerate}
\item[(C2)] for each variable $x_i \in X$ there is an element $b_i \in \mathcal B$ such that $x_i\cdot b_i \in \mathcal B$.
\end{enumerate}
Condition (C2) guarantees that the root $p$ can be directly computed from the eigenvector $u$. If $x_i\cdot b_i = b'$ and the elements $b_i$ and $b'$ are at the $r$th and $q$th positions of vector $\ve{\mathcal B}$ respectively, then $x_i(p) = u^q/u^r$, where $u^q$ and $u^r$ are the $q$th and $r$th entries of vector $u$ respectively. On the other hand, if $\mathcal B$ does not satisfy condition (C2), then additional computations may be required to derive roots.

To summarize, knowing the solving set $\mathcal B$, which additionally satisfies condition (C2), together with the Laurent polynomials from $J = \langle F\rangle$, which have the form~\eqref{eq:vectorC}, allows one to compute the roots of the system $F = 0$. The main question is, \textit{how to find the solving sets}? For this purpose we propose to use elimination templates and the incremental approach similar to that from~\cite{kukelova2008automatic}.

\section{Macaulay matrices and elimination templates}

Given a Laurent polynomial $f$, we denote by $U_f$ the support of $f$, \ie
\[
U_f = \{m \in U \cln c(f, m) \neq 0\},
\]
where $c(f, m)$ is the coefficient of $f$ at monomial $m$. Given a set of Laurent polynomials $F = \{f_1, \ldots, f_s\}$, we denote by $U_F$ the support of $F$, \ie
\[
U_F = \bigcup_{i = 1}^s U_{f_i}.
\]
Let $n = \# U_F$ be the cardinality of the finite set $U_F$. The \emph{Macaulay matrix} $M(F) \in \mathbb K^{s\times n}$ is defined as follows: its $(i, j)$th element is the coefficient $c(f_i, m_j)$ of the polynomial $f_i \in \ve{F}$ at the monomial $m_j \in U_F$, \ie ${M(F)}_{ij} = c(f_i, m_j)$. Thus,
\[
M(F)\, \ve{U_F} = 0
\]
is the vector form of the Laurent polynomial system $F = 0$.

A \emph{shift} of a polynomial $f$ is a multiple of $f$ by a monomial $m \in U$. Let $A = (A_1, \ldots, A_s)$ be an ordered $s$-tuple of finite sets of monomials $A_j \subset U$ for all $j$. We define the \emph{set of shifts} of $F$ as
\[
A\cdot F = \{m \cdot f_j \cln m \in A_j, f_j \in F\}.
\]

Let $a$ be a Laurent polynomial and $\mathcal B$ be a finite subset of Laurent monomials from $U_{A\cdot F}$ such that $U_{a\,m} \subset U_{A\cdot F}$ for each $m \in \mathcal B$. We define the two subsets 
\begin{align*}
\mathcal R &= \bigcup_{b \in U_a} \{b\,m \cln m \in \mathcal B\} \setminus \mathcal B,\\
\mathcal E &= U_{A\cdot F} \setminus (\mathcal R \cup \mathcal B).
\end{align*}
Clearly, the subsets $\mathcal B$, $\mathcal R$, $\mathcal E$ are pairwise disjoint and $U_{A\cdot F} = \mathcal E \cup \mathcal R \cup \mathcal B$.

\newcommand{\nhs}{\hspace{-8pt}}

\begin{definition}
A Macaulay matrix $M(A\cdot F)$ with columns arranged in ordered blocks $M(A \cdot F) = \begin{bmatrix}
M_{\mathcal E} & M_{\mathcal R} & M_{\mathcal B}\end{bmatrix}$ is called the \emph{elimination template} for $F$ \wrt $a$ if the reduced row echelon form of $M(A\cdot F)$ is 
\[
\widetilde M(A\cdot F) =
\nhs
\kbordermatrix{
& \mathcal E & \mathcal R & \mathcal B\\
& * & 0 & * \\
& 0 & I & \widetilde M_{\mathcal B} \\ 
& 0 & 0 & 0},
\]
where $*$ means a submatrix with arbitrary entries, $0$ is the zero matrix of a suitable size, $I$ is the identity matrix of order $\# \mathcal R$ and $\widetilde M_{\mathcal B}$ is a matrix of size $\# \mathcal R \times \#\mathcal B$.
\end{definition}

It follows from the definition that if a Macaulay matrix $M(A\cdot F)$ is an elimination template, then the set $\mathcal B$ is the solving set for $J = \langle F\rangle$. On the other hand, the action polynomial $a$, the $s$-tuple of sets $A$ and the solving set $\mathcal B$ uniquely determine the elimination template $M(A\cdot F)$ (up to reordering its rows and columns in $M_{\mathcal E}$, $M_{\mathcal R}$, $M_{\mathcal B}$).

\section{Automatic solver generator}

Our automatic solver generator consists of two main steps: (i) finding an elimination template for a set of Laurent polynomials (\templateFinder); (ii) reducing the template by removing all its unnecessary rows and columns (\templateReduction). Both steps are essentially based on the procedure that checks whether a given set of polynomials is sufficient to construct an elimination template for a given action polynomial (\templateTest). To speed up the computation, both steps are performed over a finite field of sufficiently large order. We assume that there exists a generic instance of the problem with coefficients in this field.

\subsection{Elimination template test}

For the sake of brevity, we denote the support $U_F$ of a finite set of Laurent polynomials $F$ by $\mathcal U$.

Given a Laurent polynomial $a$, we define the set of \emph{permissible monomials}~\cite{byrod2009fast} as
\[
\mathcal P = \bigcap_{b \in U_a} \{m \in \mathcal U \cln b\,m \in \mathcal U\},
\]
the set of \emph{reducible monomials} as
\[
\mathcal R = \bigcup_{b \in U_a} \{b\,m \cln m \in \mathcal P\} \setminus \mathcal P,
\]
and the set of \emph{excessive monomials} $\mathcal E$ consisting of monomials from $\mathcal U$ which are neither in $\mathcal R$ nor in $\mathcal P$, \ie
\[
\mathcal E = \mathcal U \setminus (\mathcal R \cup \mathcal P).
\]

First we set $\mathcal U_0 = \mathcal U$ and $\widetilde{\mathcal E}_0 = \varnothing$. We open the loop over the index $i$ starting with $i = 1$. At the $i$th iteration we set
\begin{align*}
\mathcal U_i &= \mathcal U_{i-1}\setminus \widetilde{\mathcal E}_{i-1},\\
\mathcal B_i &= \bigcap_{b \in U_a} \{m \in \mathcal U_i \cln b\,m \in \mathcal U_i\}.
\end{align*}
If $\mathcal B_i = \varnothing$, then the algorithm terminates with the empty set. Otherwise, we proceed
\begin{align*}
\mathcal R_i &= \bigcup_{b \in U_a} \{b\,m \cln m \in \mathcal B_i\} \setminus \mathcal B_i,\\
\mathcal E_i &= \widetilde{\mathcal E}_{i-1} \cup \mathcal U_i \setminus (\mathcal R_i \cup \mathcal B_i).
\end{align*}
Let $M$ be a Macaulay matrix corresponding to $F$ and $V$ be the related monomial vector. We reorder the columns of matrix $M$ and the entries of vector $V$ according to the partition $\mathcal E_i \cup \mathcal R_i \cup \mathcal B_i$. The resulting Macaulay matrix and the resulting monomial vector, denoted by $M_i$ and $V_i$ respectively, obey the relation $M_iV_i = MV$.

Next, let $\widetilde M_i$ be the reduced row echelon form of $M_i$ and $\widetilde F_i = \{\widetilde M_i V_i\}$ be the corresponding set of Laurent polynomials. We define the following subset of $\mathcal R_i$:
\[
\widetilde{\mathcal R}_i = \{m \in \mathcal R_i \cln m - \sum_j \gamma_j b_j \in \widetilde F_i, \gamma_j \in \mathbb K, b_j \in \mathcal B_i\}.
\]
If $\widetilde{\mathcal R}_i = \mathcal R_i$, then we set $l = i$ and terminate the loop over $i$. Otherwise, we set $\widetilde{\mathcal E}_i = \mathcal E_i \cup (\mathcal R_i \setminus \widetilde{\mathcal R}_i)$ and proceed with $i+1$.

The algorithm generates the following sequence of proper subsets
\[
\mathcal P = \mathcal B_1 \supset \mathcal B_2 \supset \ldots \supset \mathcal B_{l-1} \supset \mathcal B_l = \mathcal B.
\]
It follows that the algorithm always terminates in a finite number of steps. By the construction, the resulting subset $\mathcal B$ is either the empty set or the set satisfying condition (C1). We additionally check if $\mathcal B$ satisfies condition (C2). If so, the algorithm returns the solving set $\mathcal B$. The respective Macaulay matrix $M_l$ is the elimination template. Otherwise, the algorithm returns the empty set. The template test function is summarized in Alg.~\ref{alg:templtest}.

\newcommand{\hs}{\hspace{12pt}}

\begin{algorithm}[t]
\caption{Given a set of Laurent polynomials $F$ and an action polynomial $a$, returns either the solving set $\mathcal B$ or the empty set.}
\label{alg:templtest}

\begin{algorithmic}[1]
\STATE {\bf function} \templateTest($F, a$)
\STATE\hs $\mathcal U \gets U_F$
\STATE\hs {$M, V \gets$ Macaulay matrix and monomial vector\\\hs for $F$}
\STATE\hs $\mathcal E \gets \varnothing$
\STATE\hs $n_{\mathcal E} \gets 1$
\STATE\hs {\bf while $\#\mathcal E \neq n_{\mathcal E}$ do}
\STATE\hs\hs $\mathcal U \gets \mathcal U \setminus \mathcal E$
\STATE\hs\hs $\mathcal B \gets \bigcap_{b \in U_a} \{m \in \mathcal U \cln b\,m \in \mathcal U\}$
\STATE\hs\hs {\bf if $\#\mathcal B = 0$ then}
\STATE\hs\hs\hs {\bf return $\varnothing$}
\STATE\hs\hs {\bf end if}
\STATE\hs\hs $\mathcal R \gets \bigcup_{b \in U_a} \{b\,m \cln m \in \mathcal B\} \setminus \mathcal B$
\STATE\hs\hs $\mathcal E \gets \mathcal E \cup \mathcal U \setminus (\mathcal R \cup \mathcal B)$
\STATE\hs\hs $n_{\mathcal E} \gets \#\mathcal E$
\STATE\hs\hs $V' \gets \begin{bmatrix}\ve{\mathcal E}^\top & \ve{\mathcal R}^\top & \ve{\mathcal B}^\top \end{bmatrix}^\top$
\STATE\hs\hs $M' \gets$ a matrix such that $M'V' = MV$
\STATE\hs\hs $\widetilde M' \gets$ the reduced row echelon form of $M'$
\STATE\hs\hs $\widetilde{\mathcal R} \gets \{m \in \mathcal R \cln m - \sum_j \gamma_j b_j \in \{\widetilde M'V'\}, \gamma_j \in \mathbb K,$\\\hs\hs $b_j \in \mathcal B\}$
\STATE\hs\hs $\mathcal E \gets \mathcal E \cup (\mathcal R \setminus \widetilde{\mathcal R})$
\STATE\hs {\bf end while}
\STATE\hs {\bf if} $\mathcal B$ satisfies condition (C2) {\bf then}
\STATE\hs\hs {\bf return $\mathcal B$}
\STATE\hs {\bf else}
\STATE\hs\hs {\bf return $\varnothing$}
\STATE\hs {\bf end if}
\STATE {\bf end function}
\end{algorithmic}
\end{algorithm}

\begin{example}
\label{toy_example}

This example demonstrates the work of the template test function from Alg.~\ref{alg:templtest} on the following set of two Laurent polynomials from $\mathbb Q[x^{\pm 1}, y^{\pm 1}]$:
\[
F = \{f_1, f_2\} = \Bigl\{\frac{2y^2}{x} - 7x - 4y + 9, \frac{2x^2}{y} - 7y - 4x + 9\Bigr\}.
\]
The system $F = 0$ has the following three roots in $(\mathbb Q \setminus \{0\})^2$: $(1,1)$, $(-1,2)$, $(2,-1)$.

First, let us show that the $2\times 5$ Macaulay matrix for the initial system is an elimination template for $F$ \wrt the action monomial $a = \sfrac{x}{y}$. At the first iteration ($i = 1$) we have
\[
\begin{array}{c}
\mathcal U_1 = \{\sfrac{x^2}{y}, x, y, \sfrac{y^2}{x}, 1\},\\[5pt]
\mathcal E_1 = \{1\}, \quad \mathcal R_1 = \{\sfrac{x^2}{y}\}, \quad \mathcal B_1 = \{x, y, \sfrac{y^2}{x}\}.
\end{array}
\]
The Macaulay matrix of the initial system whose columns are arranged \wrt $\mathcal E_1 \cup \mathcal R_1 \cup \mathcal B_1$ is given by
\[
M_1 =
\kbordermatrix{& 1 & \omit & \sfrac{x^2}{y} & \omit & x & y & \sfrac{y^2}{x} \\
f_1 & 9 & \omit\vrule & 0 & \omit\vrule & -7 & -4 & 2 \\
f_2 & 9 & \omit\vrule & 2 & \omit\vrule & -4 & -7 & 0}.
\]
The reduced row echelon form of $M_1$ has the form
\[
\widetilde M_1 =
\nhs\kbordermatrix{& 1 & \omit & \sfrac{x^2}{y} & \omit & x & y & \sfrac{y^2}{x} \\
& 1 & \omit\vrule & 0 & \omit\vrule & -\sfrac{7}{9} & -\sfrac{4}{9} & \sfrac{2}{9} \\
& 0 & \omit\vrule & 1 & \omit\vrule & \sfrac{3}{2} & -\sfrac{3}{2} & -1}.
\]
The second row implies $\frac{x^2}{y} + \frac{3}{2}\, x - \frac{3}{2}\, y - \frac{y^2}{x} = 0$, \ie $\widetilde{\mathcal R}_1 = \mathcal R_1$. It follows that the matrix $M_1$ is the elimination template for $F$ \wrt $a$. The set $\mathcal B_1$ does satisfy condition (C1) but does not satisfy condition (C2): there is no element $b \in \mathcal B_1$ such that $x\cdot b \in \mathcal B_1$ or $y\cdot b \in \mathcal B_1$. Therefore, none of the two coordinates of a solution can be read off from the eigenvectors of the related action matrix. The algorithm returns the empty set.

Now let us consider the set of shifts $A\cdot F = \{f_2/x, f_2, f_1\}$ and the same action monomial $a = \sfrac{x}{y}$. At the first iteration $(i = 1)$ we have
\[
\begin{array}{c}
\mathcal U_1 = \{\sfrac{x^2}{y}, x, y, \sfrac{y^2}{x}, \sfrac{x}{y}, 1, \sfrac{y}{x}, \sfrac{1}{x}\},\\[5pt]
\mathcal E_1 = \{\sfrac{1}{x}\}, \quad \mathcal R_1 = \{\sfrac{x^2}{y}, \sfrac{x}{y}\}, \quad \mathcal B_1 = \{x, y, \sfrac{y^2}{x}, 1, \sfrac{y}{x}\}.
\end{array}
\]
The Macaulay matrix of the expanded system whose columns are arranged \wrt $\mathcal E_1 \cup \mathcal R_1 \cup \mathcal B_1$ is given by
\[
M_1 =
\kbordermatrix{& \sfrac{1}{x} & \omit & \sfrac{x^2}{y} & \sfrac{x}{y} & \omit & x & y & \sfrac{y^2}{x} & 1 & \sfrac{y}{x} \\
\frac{f_2}{x} & 9 & \omit\vrule & 0 & 2 & \omit\vrule & 0 & 0 & 0 & -4 & -7 \\
f_2 & 0 & \omit\vrule & 2 & 0 & \omit\vrule & -4 & -7 & 0 & 9 & 0 \\
f_1 & 0 & \omit\vrule & 0 & 0 & \omit\vrule & -7 & -4 & 2 & 9 & 0}.
\]
The reduced row echelon form of $M_1$ has the form
\[
\widetilde M_1 =
\nhs\kbordermatrix{& \sfrac{1}{x} & \omit & \sfrac{x^2}{y} & \sfrac{x}{y} & \omit & x & y & \sfrac{y^2}{x} & 1 & \sfrac{y}{x} \\
& 1 & \omit\vrule & 0 & \sfrac{2}{9} & \omit\vrule & 0 & 0 & 0 & -\sfrac{4}{9} & -\sfrac{7}{9} \\
& 0 & \omit\vrule & 1 & 0 & \omit\vrule & 0 & -\sfrac{33}{14} & -\sfrac{4}{7} & \sfrac{27}{14} & 0\\
& 0 & \omit\vrule & 0 & 0 & \omit\vrule & 1 & \sfrac{4}{7} & -\sfrac{2}{7} & -\sfrac{9}{7} & 0}.
\]
The last two rows imply that $\widetilde{\mathcal R}_1 = \{\sfrac{x^2}{y}\} \neq \mathcal R_1$ and hence we proceed by setting
\[
\widetilde{\mathcal E}_1 = \mathcal E_1 \cup (\mathcal R_1\setminus \widetilde{\mathcal R}_1) = \{\sfrac{x}{y}, \sfrac{1}{x}\}.
\]
At the second iteration ($i = 2$) we have
\[
\begin{array}{c}
\mathcal U_2 = \mathcal U_1 \setminus \widetilde{\mathcal E}_1 = \{\sfrac{x^2}{y}, x, y, \sfrac{y^2}{x}, 1, \sfrac{y}{x}\},\\[5pt]
\mathcal E_2 = \{\sfrac{x}{y}, \sfrac{1}{x}\}, \quad \mathcal R_2 = \{\sfrac{x^2}{y}, 1\}, \quad \mathcal B_2 = \{x, y, \sfrac{y^2}{x}, \sfrac{y}{x}\}.
\end{array}
\]
The rearranged Macaulay matrix is given by
\[
M_2 =
\kbordermatrix{& \sfrac{x}{y} & \sfrac{1}{x} & \omit & \sfrac{x^2}{y} & 1 & \omit & x & y & \sfrac{y^2}{x} & \sfrac{y}{x} \\
\frac{f_2}{x} & 2 & 9 & \omit\vrule & 0 & -4 & \omit\vrule & 0 & 0 & 0 & -7 \\
f_2 & 0 & 0 & \omit\vrule & 2 & 9 & \omit\vrule & -4 & -7 & 0 & 0 \\
f_1 & 0 & 0 & \omit\vrule & 0 & 9 & \omit\vrule & -7 & -4 & 2 & 0}.
\]
The reduced row echelon form of $M_2$ has the form
\[
\widetilde M_2 =
\nhs\kbordermatrix{& \sfrac{x}{y} & \sfrac{1}{x} & \omit & \sfrac{x^2}{y} & 1 & \omit & x & y & \sfrac{y^2}{x} & \sfrac{y}{x} \\
& 1 & \sfrac{9}{2} & \omit\vrule & 0 & 0 & \omit\vrule & -\sfrac{14}{9} & -\sfrac{8}{9} & \sfrac{4}{9} & -\sfrac{7}{2} \\
& 0 & 0 & \omit\vrule & 1 & 0 & \omit\vrule & \sfrac{3}{2} & -\sfrac{3}{2} & -1 & 0\\
& 0 & 0 & \omit\vrule & 0 & 1 & \omit\vrule & -\sfrac{7}{9} & -\sfrac{4}{9} & \sfrac{2}{9} & 0}.
\]
The last two rows imply $\widetilde{\mathcal R}_2 = \mathcal R_2$ and hence $M_2$ is the elimination template for $F$ \wrt $a = \sfrac{x}{y}$. Now the solving set $\mathcal B_2$ does satisfy condition (C2) as
\[
x\cdot \sfrac{y}{x} \in \mathcal B_2, \quad y\cdot \sfrac{y}{x} \in \mathcal B_2.
\]
Finally we note that the first two columns of matrix $M_2$, corresponding to the excessive monomials, are linearly dependent. Removing one of these columns results in the reduced elimination template of size $3\times 7$. The related action matrix is of order $4$, \ie the solver has one redundant root.
\end{example}

\subsection{Finding template}

Based on the template test function described in the previous subsection, we propose the algorithm for finding an elimination template for a given set $F$ of $s$ Laurent polynomials.

First we define the trivial $s$-tuple $A^0 = (\{1\}, \ldots, \{1\})$ such that $A^0\cdot F = F$.

We open the loop over the index $i$ starting with $i = 1$. At the $i$th iteration we expand the $s$-tuple $A^{i-1} = (A^{i-1}_1, \ldots, A^{i-1}_s)$ as follows
\[
A^i_j = A^{i-1}_j \cup \{x^{\pm 1}\cdot m \cln x \in X, m \in A^{i-1}_j\} \quad \forall j.
\]
Then we construct the set of shifts $A^i\cdot F$. For each monomial $a \in X^{-1} \cup X$, where $X^{-1} = \{x_1^{-1}, \ldots, x_k^{-1}\}$, we evaluate $\mathcal B_i = \templateTest(A^i\cdot F, a)$. If $\mathcal B_i \neq \varnothing$, then $\mathcal B_i$ is the solving set and the algorithm terminates with the data $a, A_i, \mathcal B_i$ required to construct the elimination template. Otherwise, we proceed with the $(i+1)$th iteration.

To ensure that the algorithm terminates in a finite number of steps, we limited iterations to a natural number $N$. In our experiments we found that for all (tractable) systems it is sufficient to set $N = 10$. The template finding function is summarized in Alg.~\ref{alg:gentempl}.

\begin{algorithm}[t]
\caption{Given a set of Laurent polynomials $F$ and a natural number $N$, returns either the action polynomial $a$, the $s$-tuple of sets $A$, and the solving set $\mathcal B$, or the empty set.}
\label{alg:gentempl}

\begin{algorithmic}[1]
\STATE {\bf function} \templateFinder($F$)
\STATE\hs $X \gets$ set of variables for $F$
\STATE\hs $A \gets$ $s$-tuple of $\{1\}$
\STATE\hs {\bf for $i = 1$ to $N$ do}
\STATE\hs\hs {\bf for $a$ in $X^{-1} \cup X$ do}
\STATE\hs\hs\hs $\mathcal B \gets \templateTest(A \cdot F, a)$
\STATE\hs\hs\hs {\bf if $\mathcal B \neq \varnothing$ then}
\STATE\hs\hs\hs\hs {\bf return $a, A, \mathcal B$}
\STATE\hs\hs\hs {\bf end if}
\STATE\hs\hs {\bf end for}
\STATE\hs\hs {\bf for $j = 1$ to $s$ do}
\STATE\hs\hs\hs $A_j \gets A_j \cup \{x^{\pm 1}\cdot m \cln x \in X, m \in A_j\}$
\STATE\hs\hs {\bf end for}
\STATE\hs\hs $A \gets (A_1, \ldots A_s)$
\STATE\hs {\bf end do}
\STATE\hs {{\bf return $\varnothing$} {\it $\backslash\backslash$ no template found}}
\STATE {\bf end function}
\end{algorithmic}
\end{algorithm}

\subsection{Reducing template}

In general, the template returned by Alg.~\ref{alg:gentempl} may be very large. In this subsection we propose a quite straightforward algorithm for its reduction.

Given the $s$-tuple of sets $A = (A_1, \ldots, A_s)$ and the solving set $\mathcal B$, we set $A' = A$ and $\mathcal B' = \mathcal B$. For each $j = 1, \ldots, s$ and $m_r \in A_j$ we define the intermediate $s$-tuple
\[
A'' = (A'_1, \ldots A'_{j-1}, A'_j \setminus m_r, A'_{j+1}, \ldots, A'_s).
\]
Then we evaluate $\mathcal B'' = \templateTest(A''\cdot F, a)$. It may happen that $\mathcal B'' \neq \mathcal B$. The cardinality of the solving set is allowed to decrease and is not allowed to increase while reduction. Therefore, we set $A' = A''$, $\mathcal B' = \mathcal B''$ if and only if $\mathcal B'' \neq \varnothing$ and $\#\mathcal B'' \leq \#\mathcal B$. Then we proceed with the next monomial $m_{r+1}$. If $r+1 > \# A_j$, then we proceed with $j+1$. The template reduction function is summarized in Alg.~\ref{alg:redtempl}.

The templates are also reduced by removing all linearly dependent columns corresponding to the excessive monomials as it is described in~\cite{martyushev2022optimizing}. As a result, our templates always satisfy the ``necessary condition of optimality'':
\[
\# \text{ of columns } - \# \text{ of rows } = \# \text{ of roots}.
\]

Finally, for problems that contain sparse polynomials with all constant coefficients, we applied the Schur complement reduction~\cite{martyushev2022optimizing}.

\begin{algorithm}[t]
\caption{Given a set of Laurent polynomials $F$, an action polynomial $a$, an $s$-tuple of sets $A$, and a solving set $\mathcal B$, returns the reduced $s$-tuple of sets $A'$ and the solving set $\mathcal B'$.}
\label{alg:redtempl}

\begin{algorithmic}[1]
\STATE {\bf function} \templateReduction($F, a, A, \mathcal B$)
\STATE\hs $A', \mathcal B' \gets A, \mathcal B$
\STATE\hs $d \gets \#\mathcal B$
\STATE\hs {\bf for $j = 1$ to $s$ do}
\STATE\hs\hs {\bf for $m$ in $A_j$ do}
\STATE\hs\hs\hs $A'' \gets (A'_1, \ldots A'_{j-1}, A'_j \setminus m, A'_{j+1}, \ldots, A'_s)$
\STATE\hs\hs\hs $\mathcal B'' \gets \templateTest(A''\cdot F, a)$
\STATE\hs\hs\hs {\bf if $\mathcal B'' = \varnothing$ or $\#\mathcal B'' > d$ then}
\STATE\hs\hs\hs\hs {\bf continue}
\STATE\hs\hs\hs {\bf end if}
\STATE\hs\hs\hs {\bf if $\#\mathcal B'' < d$ then}
\STATE\hs\hs\hs\hs $d \gets \#\mathcal B''$
\STATE\hs\hs\hs {\bf end if}
\STATE\hs\hs\hs $A', \mathcal B' \gets A'', \mathcal B''$
\STATE\hs\hs {\bf end for}
\STATE\hs {\bf end for}
\STATE\hs {\bf return $A', \mathcal B'$}
\STATE {\bf end function}
\end{algorithmic}
\end{algorithm}

\section{Experiments}
\label{sec:exper}

In this section, we test our solver generator on 36 minimal problems from geometric computer vision and acoustics. We compare our AG with one of the state-of-the-art AGs from~\cite{martyushev2022optimizing} (Greedy). The results are presented in Tab.~\ref{tab:res}, and we make the following remarks about them.

\begin{table*}
\centering
\scriptsize
\begin{tabular}{rlcccccccccc}
\hline\\[-6pt]
\# &
Problem & \multicolumn{5}{c}{\begin{tabular}{ccccc}\multicolumn{5}{c}{Our} \\\hline\\[-6pt] ~~~~size & ~~~~~~roots & \!\!\!time (ms) & \!\!mean & ~~~~~med.\end{tabular}} & \multicolumn{5}{c}{\begin{tabular}{ccccc}\multicolumn{5}{c}{Greedy~\cite{martyushev2022optimizing}} \\\hline\\[-6pt] ~size & ~~~roots & \!\!\!\!\!time (ms) & \!\!\!mean & ~~~~~med. \end{tabular}}\\
\hline\\[-6pt]
1 &
Rel. pose $\lambda$+$F$+$\lambda$ 8pt \cite{kukelova2008automatic} & $19 \tm 39$ & $20$ & $\bf 0.2$ & $-11.77$ & $-12.13$ & $31 \tm 47$ & $16$ & $\bf 0.2$ & $-12.19$ & $-12.59$ \\
2 &
P3.5P+focal \cite{wu2015p3} & $12 \tm 26$ & $14$ & $\bf 0.1$ & $-13.16$ & $-13.35$ & $18 \tm 28$ & $10$ & $\bf 0.1$ & $-12.32$ & $-12.51$ \\
3 &
Stitching $f\lambda$+$R$+$f\lambda$ 3pt \cite{naroditsky2011optimizing} & $6 \tm 30$ & $24$ & $\bf 0.2$ & $-12.38$ & $-12.92$ & $18 \tm 36$ & $18$ & $\bf 0.2$ & $-12.16$ & $-12.57$ \\
4 &
Abs. pose P4P+fr \cite{bujnak2010new} & $42 \tm 60$ & $18$ & $\color{blue}\bf 0.2$ & $-12.49$ & $-12.71$ & $52 \tm 68$ & $16$ & $0.3$ & $-12.40$ & $-12.64$ \\
5 &
Rel. pose $\lambda_1$+$F$+$\lambda_2$ 9pt \cite{kukelova2008automatic} & $73 \tm 97$ & $24$ & $\color{blue}\bf 0.3$ & $-10.55$ & $-10.94$ & $76 \tm 100$ & $24$ & $0.4$ & $-9.80$ & $-10.09$ \\
6 &
Rel. pose $E$+$f\lambda$ 7pt \cite{kuang2014minimal} & $34 \tm 56$ & $22$ & $\color{blue}\bf 0.2$ & $-11.52$ & $-11.90$ & $55 \tm 74$ & $19$ & $0.6$ & $-11.93$ & $-12.26$ \\
7 &
Rolling shutter pose \cite{saurer2015minimal} & $40 \tm 52$ & $12$ & $\bf 0.2$ & $-13.28$ & $-13.40$ & $47 \tm 55$ & $8$ & $\bf 0.2$ & $-13.08$ & $-13.20$ \\
8 &
Triangulation (sat. im.) \cite{zheng2015minimal} & $74 \tm 104$ & $30$ & $\color{blue}\bf 0.4$ & $-11.52$ & $-11.74$ & $87 \tm 114$ & $27$ & $0.8$ & $-11.17$ & $-11.59$ \\
9 &
Abs. pose refractive P5P \cite{haner2015absolute} & $38 \tm 58$ & $20$ & $\color{blue}\bf 0.2$ & $-12.32$ & $-12.66$ & $57 \tm 73$ & $16$ & $0.3$ & $-11.74$ & $-12.06$ \\
10 &
Abs. pose quivers \cite{kuang2013pose} & $56 \tm 80$ & $24$ & $\color{blue}\bf 0.2$ & $-11.88$ & $-12.13$ & $65 \tm 85$ & $20$ & $0.3$ & $-12.30$ & $-12.52$ \\
11 &
Unsynch. rel. pose \cite{albl2017two} & $59 \tm 79$ & $20$ & $\color{blue}\bf 0.3$ & $-14.65$ & $-14.92$ & $139 \tm 155$ & $16$ & $0.6$ & $-10.61$ & $-11.23$ \\
12 &
Optimal PnP (Cayley) \cite{nakano2015globally} & $78 \tm 131$ & $53$ & $1.1$ & $-9.67$ & $-9.94$ & $118 \tm 158$ & $40$ & $\color{blue}\bf 0.9$ & $-8.36$ & $-8.68$ \\
13 &
Rel. pose $E$+ang. 4pt v2 \cite{martyushev2020efficient} & $12 \tm 36$ & $24$ & $\bf 0.2$ & $-12.93$ & $-13.06$ & $16 \tm 36$ & $20$ & $\bf 0.2$ & $-12.99$ & $-13.09$ \\
14 &
Gen. rel. pose $E$+ang. 5pt \cite{martyushev2020efficient} & $34 \tm 79$ & $45$ & $\color{blue}\bf 0.8$ & $-11.06$ & $-11.28$ & $37 \tm 81$ & $44$ & $0.9$ & $-11.48$ & $-11.61$ \\
15 &
Rolling shutter R6P \cite{albl2015r6p} & $66 \tm 92$ & $26$ & $\color{blue}\bf 0.3$ & $-12.63$ & $-12.77$ & $120 \tm 140$ & $20$ & $0.9$ & $-12.07$ & $-12.25$ \\
16 &
Opt. pose w dir 2pt \cite{svarm2016city} & 
$87 \tm 120$ & $33$ & $\color{blue}\bf 0.5$ & $-10.35$ & $-10.58$ & $139 \tm 163$ & $24$ & $1.2$ & $-10.14$ & $-10.39$ \\
17 &
Opt. pose w dir 3pt \cite{svarm2016city} & 
$297 \tm 356$ & $59$ & $\color{blue}\bf 3.0$ & $-8.99$ & $-9.28$ & $385 \tm 433$ & $48$ & $6.8$ & $-7.80$ & $-7.97$ \\
18 &
Opt. pose w dir 4pt \cite{svarm2016city} & 
$105 \tm 138$ & $33$ & $\color{blue}\bf 0.6$ & $-11.48$ & $-11.69$ & $134 \tm 162$ & $28$ & $1.1$ & $-11.67$ & $-11.81$ \\
19 &
$L_2$ 3-view triang. (relaxed) \cite{kukelova2013fast} & 
$190 \tm 227$ & $37$ & $\color{blue}\bf 0.6$ & $-11.04$ & $-11.54$ & $217 \tm 248$ & $31$ & $1.4$ & $-11.53$ & $-11.90$ \\
20 &
Rel. pose $f\lambda$+$E$+$f\lambda$ 7pt \cite{jiang2014minimal} & $138 \tm 210$ & $72$ & $\color{blue}\bf 2.3$ & $-6.80$ & $-7.01$ & $209 \tm 277$ & $68$ & $3.0$ & $-7.27$ & $-7.46$ \\
21 &
Rel. pose $\lambda_1$+$E$+$\lambda_2$ 7pt \cite{oskarsson2021fast} & $204 \tm 289$ & $85$ & $\color{blue}\bf 4.2$ & $-5.78$ & $-5.82$ & $436 \tm 512$ & $76$ & $27.9$ & $-3.56$ & $-3.23$ \\
22 &
Gen. rel. pose 6pt \cite{stewenius2005solutions} & $78 \tm 155$ & $77$ & $\bf 2.5$ & $-8.97$ & $-9.29$ & $99 \tm 163$ & $64$ & $\bf 2.5$ & $-7.79$ & $-8.25$ \\
23 &
Rel. pose 9 lines \cite{oskarsson2004minimal,larsson2017efficient} & $1,\!610 \tm 1,\!726$ & $116$ & $\color{blue}\bf 205$ & $-1.25$ & $-1.17$ & $-$ & $-$ & $-$ & $-$ & $-$ \\
24 &
Gen. rel. pose + scale 7pt \cite{kneip2016generalized} & $130 \tm 275$ & $145$ & $9.7$ & $-3.16$ & $-2.72$ & $144 \tm 284$ & $140$ & $\color{blue}\bf 9.2$ & $-5.93$ & $-6.26$ \\
25 &
Optimal PnP \cite{zheng2013revisiting} & $120 \tm 175$ & $55$ & $\color{blue}\bf 1.5$ & $-8.68$ & $-9.04$ & $272 \tm 312$ & $40$ & $5.9$ & $-8.88$ & $-9.34$ \\
26 &
Weak PnP \cite{larsson2016uncovering} & 
$85 \tm 107$ & $22$ & $\color{blue}\bf 0.3$ & $-10.32$ & $-10.87$ & $108\tm 124$ & $16$ & $0.6$ & $-12.36$ & $-12.72$ \\
27 &
Weak PnP ($2\tm 2$ sym) \cite{larsson2016uncovering} & 
$10 \tm 20$ & $10^*$ & $\color{blue}\bf 0.1$ & $-12.67$ & $-13.16$ & $18 \tm 34$ & $16$ & $0.2$ & $-10.62$ & $-12.98$ \\
28 &
Refractive P6P+focal \cite{haner2015absolute} & $40 \tm 61$ & $21^*$ & $\color{blue}\bf 0.2$ & $-10.69$ & $-11.01$ & $126 \tm 162$ & $36$ & $0.8$ & $-10.50$ & $-10.88$ \\
29 &
Rel. pose $E$+$fuv$+ang. \cite{martyushev2018self} & $26 \tm 33^\dag$ & $7$ & $\color{blue}\bf 0.1$ & $-12.20$ & $-12.66$ & $40 \tm 46$ & $6$ & $0.2$ & $-11.77$ & $-12.05$ \\
30 &
Vanishing point est. \cite{mirzaei2011optimal,larsson2017polynomial} & $136 \tm 194^\dag$ & $58$ & $\color{blue}\bf 2.1$ & $-4.69$ & $-4.84$ & $343 \tm 383$ & $40$ & $13.8$ & $-5.67$ & $-5.99$ \\
31 &
Time-of-Arrival (4,6) \cite{kuang2013complete,larsson2017polynomial} & $427 \tm 475^\dag$ & $48$ & $\color{blue}\bf 5.2$ & $-5.24$ & $-5.45$ & $863 \tm 901$ & $38$ & $17.1$ & $-2.23$ & $-1.63$ \\
32 &
Time-of-Arrival (5,5) \cite{kuang2013complete,larsson2017polynomial} & $772 \tm 832^\dag$ & $60$ & $\color{blue}\bf 17.8$ & $-5.49$ & $-5.60$ & $-$ & $-$ & $-$ & $-$ & $-$ \\
33 &
Semi-gen. rel. pose H51f \cite{bhayani2023hybsemigenpose} & $740 \tm 811^\dag$ & $71$ & $\color{blue}\bf 43.4$ & $-4.47$ & $-4.70$ & $-$ & $-$ & $-$ & $-$ & $-$ \\
34 &
Semi-gen. rel. pose H32f \cite{bhayani2023hybsemigenpose} & $174 \tm 222^\dag$ & $48$ & $\color{blue}\bf 2.8$ & $-9.56$ & $-9.82$ & $258 \tm 284$ & $26$ & $5.7$ & $-8.64$ & $-9.00$ \\
35 &
Opt. 3-view triang. \cite{stewenius2005hard,larsson2017polynomial} & 
$69 \tm 127^\ddag$ & $58$ & $\color{blue}\bf 1.1$ & $-8.09$ & $-8.47$ & $401 \tm 448$ & $47$ & $5.7$ & $-2.74$ & $-1.58$ \\
36 &
Semi-gen. rel. pose H13f (depth) & 
$115 \tm 134^\ddag$ & $19$ & $\color{blue}\bf 1.1$ & $-9.94$ & $-10.27$ & $-$ & $-$ & $-$ & $-$ & $-$ \\
\hline\\
\end{tabular}
\caption{A comparison of the elimination templates for our test minimal problems. The minimal runtimes are shown in bold, the runtimes which are smaller than the state-of-the-art are shown in blue bold. ($*$): due to the symmetry the number of roots is less than the degree of the ideal; ($\dag$): template constructed for the positive-dimensional formulation; ($\ddag$): template constructed for the Laurent polynomial formulation; ($-$): failed to construct template in a reasonable time}
\label{tab:res}
\end{table*}

\noindent{\bf 1.} The experiments were performed on a system with Intel(R) Core(TM) i5-1155G7 @ 2.5~GHz and 8~GB of RAM.

\medskip

\noindent{\bf 2.} In general, the size of a template alone is not an appropriate measure of the efficiency of the corresponding solver. For example, the 5-point absolute pose estimation problem for a known refractive plane (Problem \#9) has the templates of sizes $38\times 58$ and $57\times 73$. The first template is smaller but is followed by the eigendecomposition of a $20\times 20$ matrix. On the other hand, the second template is larger but requires the eigendecomposition of a smaller matrix of size $16\times 16$. At first glance, it is unclear which of these two templates would provide a faster solver. Therefore, to compare the efficiency of the solvers, we reported the template size, the number of roots and the average runtime of the corresponding Matlab~\cite{matlab2019} implementation. The reported times include the derivation of the action matrix and its eigendecomposition and do not include the construction of the coefficient matrix.

\medskip

\noindent{\bf 3.} The numerical error is defined as follows. Let the Laurent polynomial system $F = 0$ be written in the form $M(F) Z = 0$, where $M(F)$ and $Z = \ve{U_F}$ are the Macaulay matrix and monomial vector respectively. The matrix $M(F)$ is normalized so that each its row has unit length. Let $d_0$ be the number of roots to $F = 0$ and $d \geq d_0$ be the number of roots returned by our solver, \ie there are $d - d_0$ false roots. Let $Z_i$ be the monomial vector $Z$ evaluated at the $i$th (possibly false) root. We compute $d$ values $\epsilon_i = \Bigl\|M(F)\frac{Z_i}{\|Z_i\|_2}\Bigr\|_2$, where $\|\cdot\|_2$ is the Frobenius norm. Then the numerical error for our solvers is measured by the value
$
\frac{1}{2}\log_{10} \sum_i \epsilon_i^2,
$
where the sum is taken over $d_0$ smallest values of $\epsilon_i$.

\medskip

\noindent{\bf 4.} The hard minimal problem of relative pose estimation from $9$ lines in $3$ uncalibrated images (Problem \#23) was first addressed in~\cite{oskarsson2004minimal} where, using the homotopy continuation method, it was shown that the problem has $36$ solutions. In~\cite{larsson2017efficient}, the authors proposed an efficient formulation of the problem consisting of $21$ polynomials in $14$ variables and first attempted to propose an eigenvalue solver for this problem by constructing a giant elimination template of size $16,278\times 13,735$. We started with exactly the same formulation as in~\cite{larsson2017efficient}. By applying the G--J elimination on the initial coefficient matrix, we excluded $4$ variables resulting in the formulation consisting of $17$ polynomials in $10$ variables. Our generator found the template of size $2,163\times 2,616$ with $116$ roots in approximately 20 minutes. Then it was reduced to the reported size in approximately 13 hours.

\medskip

\noindent{\bf 5.} Problems \#13, \#14, \#16, \#17, \#18, \#30 contain sparse polynomials with all (or all but one) constant coefficients. We additionally reduced the templates for these problems by the Schur complement reduction, see~\cite{martyushev2022optimizing} for details.

\medskip

\noindent{\bf 6.} The 2-fold symmetries in the formulations of Problems \#25 and \#26 were uncovered manually by changing variables. On the other hand, our generator automatically uncovered the partial symmetries for Problems \#27 and \#28 by constructing the solving set of cardinality less than the degree of the related ideal.

\medskip

\noindent{\bf 7.} The AG from~\cite{martyushev2022optimizing} applies only to zero-dimensional ideals. Therefore, to apply it to Problems \#29--\#36, we saturated the positive-dimensional components in their formulations either by the Rabinowitsch trick~\cite{rabinowitsch1930hilbertschen}, or by a cascade of G--J eliminations as in~\cite{martyushev2018self}. The remaining problems were compared using the same formulations.

\medskip

\noindent{\bf 8.} The Maple~\cite{maple2020maplesoft} implementation of the new AG, as well as the Matlab~\cite{matlab2019} solvers for all the minimal problems from Tab.~\ref{tab:res}, are made publicly available at \href{https://github.com/martyushev/EliminationTemplates}{https://github.com/martyushev/EliminationTemplates}.

\subsection{Optimal 3-view triangulation}

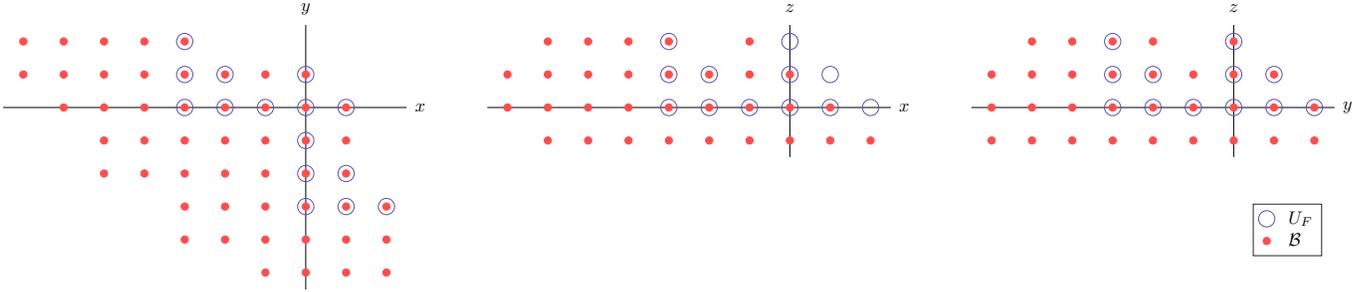
\begin{figure*}
\centering
\resizebox{\textwidth}{!}{
%
%
\definecolor{mycolor1}{rgb}{0.40000,0.30000,0.70000}%
\begin{tikzpicture}

\begin{axis}[%
width=11.028in,
height=3.151in,
at={(1.85in,1.934in)},
xmin=-7.5,
xmax=26.2,
ymin=-5.5,
ymax=4.5,
axis line style={draw=none},
ticks=none,
legend style={at={(0.97,0.1)}, anchor=south east, legend cell align=left, align=left, draw=white!15!black}
]
\addplot[only marks, mark=o, mark options={}, mark size=4.5pt, draw=mycolor1] table[row sep=crcr]{%
x	y\\
2	-3\\
1	-0\\
1	-2\\
1	-3\\
-0	1\\
-0	-0\\
-0	-1\\
-0	-2\\
-0	-3\\
-1	-0\\
-2	1\\
-2	-0\\
-3	2\\
-3	1\\
-3	-0\\
14	-0\\
13	1\\
13	-0\\
12	2\\
12	1\\
12	-0\\
11	-0\\
10	1\\
10	-0\\
9	2\\
9	1\\
9	-0\\
25	-0\\
24	1\\
24	-0\\
23	2\\
23	1\\
23	-0\\
22	-0\\
21	1\\
21	-0\\
20	2\\
20	1\\
20	-0\\
};
\addlegendentry{{}~~$U_F$}

\addplot[only marks, mark=*, mark options={}, mark size=2.0pt, color=white!30!red, fill=white!30!red] table[row sep=crcr]{%
x	y\\
2	-3\\
2	-4\\
2	-5\\
2	-6\\
1	-0\\
1	-1\\
1	-2\\
1	-3\\
1	-4\\
1	-5\\
1	-6\\
-0	1\\
-0	-0\\
-0	-1\\
-0	-2\\
-0	-3\\
-0	-4\\
-0	-5\\
-0	-6\\
-1	1\\
-1	-0\\
-1	-1\\
-1	-2\\
-1	-3\\
-1	-4\\
-1	-5\\
-2	1\\
-2	-0\\
-2	-1\\
-2	-2\\
-2	-3\\
-2	-4\\
-3	2\\
-3	1\\
-3	-0\\
-3	-1\\
-3	-2\\
-3	-3\\
-3	-4\\
-4	2\\
-4	1\\
-4	-0\\
-4	-1\\
-4	-2\\
-5	2\\
-5	1\\
-5	-0\\
-5	-1\\
-5	-2\\
-6	2\\
-6	1\\
-6	-0\\
-7	2\\
-7	1\\
14	-1\\
13	-0\\
13	-1\\
12	1\\
12	-0\\
12	-1\\
11	2\\
11	1\\
11	-0\\
11	-1\\
10	1\\
10	-0\\
10	-1\\
9	2\\
9	1\\
9	-0\\
9	-1\\
8	2\\
8	1\\
8	-0\\
8	-1\\
7	2\\
7	1\\
7	-0\\
7	-1\\
6	2\\
6	1\\
6	-0\\
6	-1\\
5	1\\
5	-0\\
25	-0\\
25	-1\\
24	1\\
24	-0\\
24	-1\\
23	2\\
23	1\\
23	-0\\
23	-1\\
22	1\\
22	-0\\
22	-1\\
21	2\\
21	1\\
21	-0\\
21	-1\\
20	2\\
20	1\\
20	-0\\
20	-1\\
19	2\\
19	1\\
19	-0\\
19	-1\\
18	2\\
18	1\\
18	-0\\
18	-1\\
17	1\\
17	-0\\
17	-1\\
};
\addlegendentry{{}~~$\mathcal B$}

\addplot [color=black]
  table[row sep=crcr]{%
-7.5	0\\
2.5	0\\
};
\addplot [color=black]
  table[row sep=crcr]{%
4.5	0\\
14.5	0\\
};
\addplot [color=black]
  table[row sep=crcr]{%
16.5	0\\
25.5	0\\
};
\addplot [color=black]
  table[row sep=crcr]{%
0	-6.5\\
0	2.5\\
};
\addplot [color=black]
  table[row sep=crcr]{%
12	-1.5\\
12	2.5\\
};
\addplot [color=black]
  table[row sep=crcr]{%
23	-1.5\\
23	2.5\\
};
\node[right, align=left, inner sep=0]
at (axis cs:2.7,0) {$x$};
\node[centered, align=left, inner sep=0]
at (axis cs:0,3) {$y$};
\node[right, align=left, inner sep=0]
at (axis cs:14.7,0) {$x$};
\node[centered, align=left, inner sep=0]
at (axis cs:12,3) {$z$};
\node[right, align=left, inner sep=0]
at (axis cs:25.7,0) {$y$};
\node[centered, align=left, inner sep=0]
at (axis cs:23,3) {$z$};
\end{axis}

\end{tikzpicture}%
}
\caption{The support $U_F$ of the Laurent polynomial system for the problem of optimal 3-view triangulation and the related solving set $\mathcal B$}
\label{fig:pts}
\end{figure*}

\begin{figure}
\centering
\resizebox{0.45\textwidth}{!}{
\input{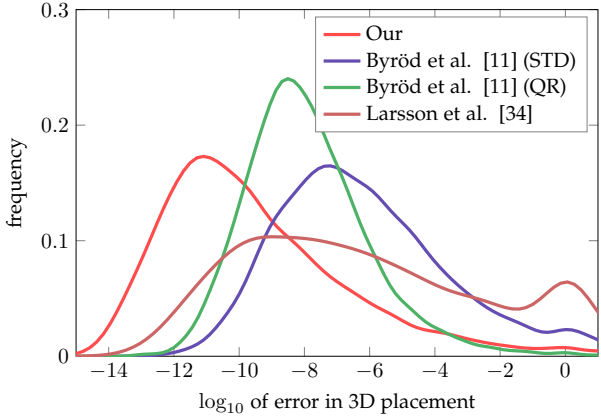}
}
\caption{The distribution of the error in 3D placement for the problem of optimal 3-view triangulation}
\label{fig:3v_triang}
\end{figure}

The optimal 3-view triangulation problem, first addressed in~\cite{stewenius2005hard}, is formulated as follows. Given three projective camera matrices $P_1$, $P_2$, $P_3$ and image point correspondences $x_1 \leftrightarrow x_2 \leftrightarrow x_3$, find the space point $X^*$ so that the reprojection error is minimized. That is
\[
X^* = \arg\min_X \sum_{i = 1}^3 \left(\frac{P^1_i X}{P^3_i X} - x^1_i\right)^2 + \left(\frac{P^2_i X}{P^3_i X} - x^2_i\right)^2,
\]
where $X = \begin{bmatrix}x & y & z & 1\end{bmatrix}^\top$, $P^j_i$ is the $j$th row of $P_i$ and $x^j_i$ is the $j$th entry of $x_i$. By choosing an appropriate projective coordinate frame, we can assume that
\begin{align*}
P^3_1 &= \begin{bmatrix}1 & 0 & 0 & 0\end{bmatrix},\\
P^3_2 &= \begin{bmatrix}0 & 1 & 0 & 0\end{bmatrix},\\
P^3_3 &= \begin{bmatrix}0 & 0 & 0 & 1\end{bmatrix},
\end{align*}
\ie the image plane of the third camera is the plane at infinity. Such parametrization, proposed in~\cite{larsson2017polynomial}, leads to smaller templates compared to $P^3_3 = \begin{bmatrix}0 & 0 & 1 & 0\end{bmatrix}$ proposed in~\cite{stewenius2005hard}.

The optimal solution is one of the $47$ stationary points which are found as roots of a system of three Laurent polynomial equations in three variables $x$, $y$, $z$. Unlike previous work, our generator is able to work directly with the Laurent polynomial formulation.

The problem has been extensively studied~\cite{stewenius2005hard,byrod2007fast,byrod2008column,larsson2017polynomial}. The solvers from~\cite{byrod2008column,larsson2017polynomial} are currently the state-of-the-art.

In Fig.~\ref{fig:pts}, we show the support $U_F$ of the initial system as well as the solving set $\mathcal B$ with $\# \mathcal B = 58$. The related elimination template is of size $69\times 127$, \cf Tab.~\ref{tab:res}, Problem \#35.

\newcommand{\px}{{\rm px}}

We tested the new solver on synthetic scenes. We modeled a 3D point $X$ lying in a cube with edge of length $1$ centered at the coordinate origin. The point is viewed by three cameras. The centers $c_i$ (here and below $i = 1,2,3$) of the cameras randomly lie on a sphere of radius $1$ also centered at the origin. The three rotation matrices $R_i$ are chosen randomly and the calibration matrices $K_i$ all have the focal length and the principal point approximately $1,000\px$ and $(500\px, 500\px)$ respectively. The initial data for our solver are the three camera matrices $P_i = K_i \begin{bmatrix}R_i & -R_ic_i\end{bmatrix}$ and the projections $x_i = P_i X$ normalized so that $x_i^3 = 1$.

We tested the numerical accuracy of our solver by constructing the distribution of the errors in 3D placement on noise-free image data. We kept the real roots, including false ones, and then picked out the unique root by calculating the reprojection errors. The 3D placement error distributions for $10$K trials are compared in Fig.~\ref{fig:3v_triang}.

The speed and the failure rate of the solvers are compared in Tab.~\ref{tab:3v_triang}.

\begin{table}
\centering
\footnotesize
\begin{tabular}{lrrrr}
\hline\\[-6pt]
Solver & Our & \cite{byrod2008column} (STD) & \cite{byrod2008column} (QR) & \cite{larsson2017polynomial} \\
\hline\\[-6pt]
Time/call & $1.34$ms & $1.54$ms & $2.07$ms & $1.56$ms \\
Relative time & $1$ & $1.15$ & $1.54$ & $1.16$ \\
Fail (error $> 1$) & $1.11\%$ & $3.09\%$ & $0.29\%$ & $8.91\%$ \\
Fail (error $> 0.1$) & $1.74\%$ & $5.01\%$ & $0.54\%$ & $13.97\%$ \\
\hline\\[-3pt]
\end{tabular}
\caption{\vspace{-20pt}}
\label{tab:3v_triang}
\end{table}

\subsection{Semi-generalized hybrid relative pose: \texorpdfstring{$\mathbf{H}13f$}{Lg}}

Consider the problem of registering a partially calibrated pinhole camera $\cam{P}$ (with unknown focal length $f$) \wrt a generalized camera $\cam{G}$ from a hybrid set of point correspondences, \ie one 2D-2D correspondence $p_1 \leftrightarrow (q_{11}, t_{g_1})$ and three 2D-3D correspondences $p_j \leftrightarrow X_j, \ j=1, \ldots, 3$. The generalized camera $\cam{G}$ is considered as a set of multiple pinhole cameras, $\{\cam{G}_i\}$, which have been registered \wrt a global coordinate frame.

The goal is to estimate the relative pose, \ie the rotation $R$ and the translation $T$, required to align the coordinate frame of $\cam{P}$ \wrt to the global coordinate frame, as well as its focal length $f$. This problem was studied in~\cite{bhayani2023hybsemigenpose} using a homography matrix-based formulation, leading to a system of two degree-$3$, one degree-$4$ and three degree-$8$ polynomials in three variables. Using~\cite{larsson2018beyond} led to a minimal solver with a template of size $70 \times 82$ with $12$ roots. However, the polynomial coefficients are quite complicated, resulting in an inefficient execution time of $55$~ms.

Instead, we generated a more efficient solver using a depth-based problem formulation. The pose and the focal length are constrained via the following equations:
\begin{equation}
\label{eq:h13f}
\begin{split}
\alpha_{1} R K^{-1} p_1 + T &= \beta_{11} q_{11} + t_{g_1},\\[5pt]
\alpha_{j} R K^{-1} p_j + T &= X_j, \quad j = 2, \ldots, 4,
\end{split}
\end{equation}
where $K = \text{diag}([f, f, 1])$ is the calibration matrix for $\cam{P}$, $\alpha_j$ and $\beta_{ij}$ denote the depths of the $j$th 3D point in the coordinate frames of $\cam{P}$ and $\cam{G}_i$ respectively. Without loss of generality, we transform the coordinate frame of $\cam{G}$ such that its origin coincides with the camera center of $\cam{G}_1$, \ie $t_{g_1} = \begin{bmatrix} 0 & 0 & 0 \end{bmatrix}^\top$. For the sake of brevity, assume $X_1 = \beta_{11} q_{11}$ in Eq.~\eqref{eq:h13f}. Eliminating $T$ from Eq.~\eqref{eq:h13f} gives the following equations:
\begin{equation}
\label{eq:h13f_elim_T}
R K^{-1} (\alpha_{i_1} p_{i_1} - \alpha_{i_2} p_{i_2}) = X_{i_1} - X_{i_2},
\end{equation}
where $ 1 \! \leq \! i_1 \! \leq 4, i_1 \! <\! i_2\! \leq\! 4 $. For the sake of brevity, assume $Y_{i_1,i_2} = \alpha_{i_1} p_{i_1} - \alpha_{i_2} p_{i_2}$ and $X_{i_1,i_2} = X_{i_1} - X_{i_2}$. Eliminating $R$ from Eq.~\eqref{eq:h13f_elim_T} yields
\begin{equation}
\label{eq:h13f_elim_RT}
\begin{split}
\|K^{-1} Y_{i_1,i_2}\|_2^2 &= \|X_{i_1,i_2}\|_2^2,\\[5pt]
Y_{i_1,i_2}^\top K^{-2} Y_{i_3,i_4} &= X_{i_1,i_2}^\top X_{i_3,i_4},
\end{split}
\end{equation}
where $1\! \leq\! i_1,\! i_3\! \leq\! 4, i_1\! <\! i_2\! \leq\! 4, i_3\! <\! i_4\! \leq\! 4, (i_1,i_2)\! \neq\! (i_3,i_4)$.
Equation~\eqref{eq:h13f_elim_RT} denotes the depth formulation for the minimal problem and consists of $20$ Laurent polynomials in $6$ variables viz., $\alpha_1, \alpha_2, \alpha_3, \alpha_4, \beta_{11}, f$. The depth formulation tends to induce polynomials in more variables, but with much simpler coefficients, than those resulting from the homography formulation. The effect is primarily observed in the execution times of the minimal solvers based on the proposed formulation versus the homography-based formulation.

Table~\ref{tab:hyb13f_time_failper_comparison} (\textbf{Row 1}) shows the average time taken/call, measured for both the proposed and the SOTA homography-based solvers.

\begin{table}
\centering
\footnotesize
\begin{tabular}{lrrr}
\hline\\[-6pt]
Solver & Our & \cite{bhayani2023hybsemigenpose} (GB~\cite{larsson2017efficient}) & \cite{bhayani2023hybsemigenpose} (Res~\cite{bhayani2021computing}) \\
\hline\\[-6pt]
Time/call & $1.10$ms & $27.41$ms & $33.00$ms \\
Relative time & $1$ & $24.74$ & $30$ \\
Fail (error $> 1$) & $0.02\%$ & $0.68\%$ & $0.06\%$ \\
Fail (error $> 0.1$) & $0.96\%$ & $5.78\%$ & $0.2\%$ \\
\hline\\[-3pt]
\end{tabular}
\caption{\vspace{-20pt}}
\label{tab:hyb13f_time_failper_comparison}
\end{table}

We also evaluated the numerical performance of the proposed depth-based solver for synthetic scenes. For this purpose, we generated $5$K 3D scenes with known ground truth parameters. In each scene, the 3D points were randomly distributed within a cube of dimensions $10 \times 10 \times 10$ units. Note that for the $\mathbf{H}13f$ case, there is only one 2D-2D point correspondence. Therefore, each 3D point was projected into two pinhole cameras with realistic focal lengths. One camera acts as a query camera, $\cam{P}$, which has to be registered, while the other camera represents the generalized camera $\cam{G}$ (consisting of only one pinhole camera). The orientations and positions of the cameras were randomly chosen so that they looked at the origin from a random distance of $15$ to $25$ units from the scene. The simulated images had a resolution of $1,000\times 1,000$~pixels. The failure rate for focal length estimation is reported in Tab.~\ref{tab:hyb13f_time_failper_comparison} (\textbf{Row 3} and \textbf{Row 4}). Note that the proposed solver has a lower or comparable failure rate than the SOTA homography-based minimal solvers~\cite{bhayani2023hybsemigenpose} generated using the Gr\"obner basis~\cite{larsson2017efficient} and the resultant~\cite{bhayani2021computing}. At the same time, the proposed solver is $20$ to $30$ times faster than the two SOTA solvers.

\begin{figure}
\centering
\resizebox{0.45\textwidth}{!}{
\input{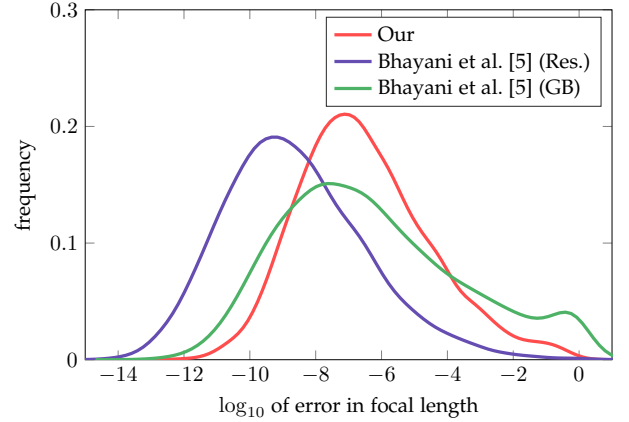}
}
\caption{The distribution of the error in focal length for the $\mathbf{H}13f$ minimal problem}
\label{fig:hyb13f_hist_f}
\end{figure}

\begin{figure}
\centering
\resizebox{0.45\textwidth}{!}{
\input{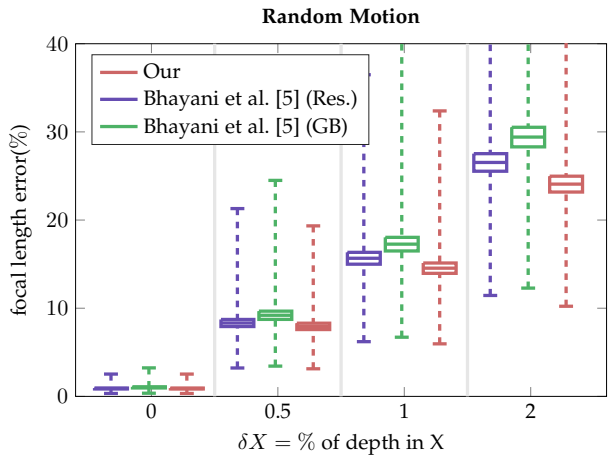}
}
\caption{A boxplot depicting the error in focal length estimates for the problem $\mathbf{H}13f$, in the presence of varying noise in the 3D points and $0.5\px$ image noise}
\label{fig:hyb13f_boxplot_f}
\end{figure}

We also evaluated the solver performance in the presence of noisy scene points by introducing Gaussian noise into the coordinates of the 3D points sampled in the synthetic scene. The standard deviation of the noise was varied as a percentage of their depths, to simulate the different quality of the keypoints used to triangulate these 3D points. We also introduced $0.5\px$ image noise to simulate noisy feature detection. For such a scene setup, we evaluated the stability of the proposed depth-based minimal solver against the SOTA homography-based minimal solvers in~\cite{bhayani2023hybsemigenpose} using the methods based on Gr\"obner bases and resultants. Figure~\ref{fig:hyb13f_boxplot_f} shows the error in focal length estimated by the solvers. Here, the box plots show the ${25}\%$ to ${75}\%$ quantiles as boxes with a horizontal line for the median. We note that our proposed depth-based solver has fewer errors, even with increasing noise in the 3D points, compared to the homography-based solvers from~\cite{bhayani2023hybsemigenpose}.

\subsection{Time-of-Arrival self-calibration}

The Time-of-Arrival (ToA) $(m,n)$ problem is formulated as follows. Given $m\times n$ distance measurements $d_{ij}$, $i = 1, \ldots, m$, $j = 1, \ldots, n$, find $m$ points $s_i$ (senders) and $n$ points $r_j$ (receivers) in 3-space such that $d(s_i, r_j) = d_{ij}$ for all $i,j$. Here $d(x, y) = \|x - y\|_2$ is the distance function. All the points (senders and receivers) are assumed to be in general position in space. Clearly, any solution to the ToA problem can be only found up to an arbitrary Euclidean isometry.

In the real world, the ToA problem arises from measuring the absolute travel times from unknown senders (\eg speakers) to unknown receivers (\eg microphones). If the signal speed is known, then the distances between the senders and receivers are also known, and we arrive at the ToA problem.

The ToA $(4,6)$ and $(5,5)$ problems are minimal and have up to $38$ and $42$ solutions respectively. These problems have been studied in papers~\cite{kuang2013complete,larsson2017polynomial,larsson2020upgrade}. The solvers from~\cite{larsson2020upgrade} are currently the state-of-the-art.

We used the ToA problem parametrization proposed in~\cite{kuang2013complete}. The $(4,6)$ problem is formulated as a system of four polynomials of degree $3$ and one of degree $4$ in $5$ unknowns. The related affine variety is the union of two subvarieties of dimensions $1$ and $0$. The $1$-dimensional component consists of superfluous roots that have no feasible interpretation, while the $0$-dimensional component consists of $38$ feasible (complex) solutions to the problem.

Similarly, the $(5,5)$ problem is formulated as a system of five polynomials of degree $3$ and one of degree $4$ in $6$ unknowns. The related variety is the union of a $2$-dimensional ``superfluous'' subvariety and a $0$-dimensional component consisting of $42$ complex roots.

Our generator automatically found the redundant solving sets of cardinality $48$ for the $(4,6)$ problem and of cardinality $60$ for the $(5,5)$ problem. The respective elimination templates are of size $427\times 475$ and $772\times 832$, see Tab.~\ref{tab:res}, Problems~\#31 and~\#32.

We tested the new solvers on synthetic scenes. We modeled $m$ senders and $n$ receivers uniformly distributed in a cube with edge of length $1$. The ground truth positions of the receivers and senders are $3$-vectors $s_i$ and $r_j$, respectively. The initial data for our solvers are the $m\times n$ distances $d(s_i, r_j)$ for all $i = 1, \ldots, m$, $j = 1, \ldots, n$.

\begin{figure}
\centering
\resizebox{0.45\textwidth}{!}{
\input{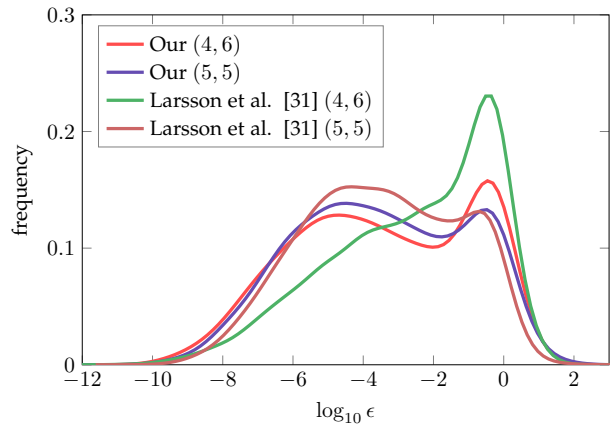}
}
\caption{The error distribution for the Time-of-Arrival $(4,6)$ and $(5,5)$ problems}
\label{fig:toa}
\end{figure}

We tested the numerical stability of the solvers on noise-free data by measuring the following error:
\begin{multline*}
\epsilon = \min\biggl(\sum_{k > i} (d(s_i, s_k) - d(\hat s_i, \hat s_k) )^2 \\[-5pt]+ \sum_{l > j} (d(r_j, r_l) - d(\hat r_j, \hat r_l) )^2\biggr)^{1/2},
\end{multline*}
where $\hat s_i$ and $\hat r_j$ are the estimated positions of the senders and receivers and the minimum is taken over all real roots. The results are presented in Fig.~\ref{fig:toa}.

The speed and the failure rate of the solvers are compared in Tab.~\ref{tab:toa}.

\begin{table}
\centering
\footnotesize
\begin{tabular}{lrrrr}
\hline\\[-6pt]
Solver & Our $(4,6)$ & \cite{larsson2020upgrade} $(4,6)$ & Our $(5,5)$ & \cite{larsson2020upgrade} $(5,5)$ \\
\hline\\[-6pt]
Time/call & $6.75$ms & $8.97$ms & $18.68$ms & $33.55$ms \\
Relative time & $1$ & $1.33$ & $1$ & $1.80$ \\
Fail (no sol.) & $5.1\%$ & $9.9\%$ & $3.3\%$ & $2.7\%$ \\
Fail ($\epsilon > 1$) & $10.8\%$ & $17.1\%$ & $9.1\%$ & $5.8\%$ \\
Fail ($\epsilon > 0.1$) & $28.7\%$ & $40.1\%$ & $23.4\%$ & $19.3\%$ \\
\hline\\[-3pt]
\end{tabular}
\caption{\vspace{-20pt}}
\label{tab:toa}
\end{table}

\section{Conclusion}

In this paper, we have proposed a new algorithm for automatically generating small and stable elimination templates for solving Laurent polynomial systems. The proposed automatic generator is flexible, versatile, and easy-to-use. It is applicable to polynomial ideals with positive-dimensional components. It is also useful for automatically uncovering the partial $p$-fold symmetries, thereby leading to smaller templates. Using the proposed automatic generator, we have been able to generate state-of-the-art elimination templates for many minimal problems, leading to substantial improvement in the solver performance.

\ifCLASSOPTIONcompsoc
  \section*{Acknowledgments}
\else
  \section*{Acknowledgment}
\fi
Snehal Bhayani has been supported by a grant from the Finnish Foundation for Technology Promotion. T. Pajdla was supported by EU H2020 SPRING No.~871245 project.

\ifCLASSOPTIONcaptionsoff
  \newpage
\fi

\bibliographystyle{amsplain}
\bibliography{biblio}

\end{document}